\documentclass[11pt,a4paper]{article}
\usepackage[hyperref]{acl2017}
\usepackage{times}
\usepackage{latexsym}

\usepackage{tabularx}
\usepackage{amsmath,amssymb,amsthm}
\usepackage{graphicx}
\usepackage{subcaption}
\usepackage{listings}
\usepackage{upquote}
\usepackage{wrapfig}
\usepackage{lipsum}
\usepackage{array}
\usepackage{multicol}
\usepackage{multirow}
\usepackage{url}

\aclfinalcopy % Uncomment this line for the final submission
\title{Rethinking Skip-thought: A Neighborhood based Approach}

\author{Shuai Tang$^1$, Hailin Jin$^2$, Chen Fang$^2$, Zhaowen Wang$^2$,  Virginia R. de Sa$^1$\\
  $^1$Department of Cognitive Science, UC San Diego, La Jolla CA 92093, USA \\
  $^2$Adobe Research, 345 Park Ave., San Jose CA 95110, USA \\
  {\tt \{shuaitang93,desa\}@ucsd.edu, \{hljin,cfang,zhawang\}@adobe.com }\\ }

\date{}

\begin{document}
\maketitle
\begin{abstract}
We study the skip-thought model proposed by \citet{Kiros2015SkipThoughtV} with neighborhood information as weak supervision. More specifically, we propose a skip-thought neighbor model to consider the adjacent sentences as a neighborhood. We train our skip-thought neighbor model on a large corpus with continuous sentences, and then evaluate the trained model on 7 tasks, which include semantic relatedness, paraphrase detection, and classification benchmarks. Both quantitative comparison and qualitative investigation are conducted. We empirically show that, our skip-thought neighbor model performs as well as the skip-thought model on evaluation tasks. In addition, we found that, incorporating an autoencoder path in our model didn't aid our model to perform better, while it hurts the performance of the skip-thought model. 
\end{abstract}

\section{Introduction}
We are interested in learning distributed sentence representation in an unsupervised fashion. Previously, the skip-thought model was introduced by \citet{Kiros2015SkipThoughtV}, which learns to explore the semantic continuity within adjacent sentences \cite{harris1954distributional} as supervision for learning a generic sentence encoder. The skip-thought model encodes the current sentence and then decodes the previous sentence and the next one, instead of itself. Two independent decoders were applied, since intuitively, the previous sentence and the next sentence should be drawn from 2 different conditional distributions, respectively.

By posing a hypothesis that the adjacent sentences provide the same neighborhood information for learning sentence representation, we first drop one of the 2 decoders, and use only one decoder to reconstruct the surrounding 2 sentences at the same time. The empirical results show that our skip-thought neighbor model performs as well as the skip-thought model on 7 evaluation tasks. Then, inspired by \citet{Hill2016LearningDR}, as they tested the effect of incorporating an autoencoder branch in their proposed FastSent model, we also conduct experiments to explore reconstructing the input sentence itself as well in both our skip-thought neighbor model and the skip-thought model. From the results, we can tell that our model didn't benefit from the autoencoder path, while reconstructing the input sentence hurts the performance of the skip-thought model. Furthermore, we conduct an interesting experiment on only decoding the next sentence without the previous sentence, and it gave us the best results among all the models. Model details will be discussed in Section \ref{approach}. 

\section{Related Work}
Distributional sentence representation learning involves learning word representations and the compositionality of the words within the given sentence. Previously, \citet{Mikolov2013DistributedRO} proposed a method for distributed representation learning for words by predicting surrounding words, and empirically showed that the additive composition of the learned word representations successfully captures contextual information of phrases and sentences. Similarly, \citet{Le2014DistributedRO} proposed a method that learns a fixed-dimension vector for each sentence, by predicting the words within the given sentence. However, after training, the representation for a new sentence is hard to derive, since it requires optimizing the sentence representation towards an objective.

% With the development of deep learning techniques, recurrent neural networks (RNNs) show encouraging results on natural language processing (NLP) tasks, and become the dominant methods in processing sequential data. Long short term memory (LSTM) \cite{Hochreiter1997LongSM} and gated recurrent unit (GRU) \cite{Cho2014LearningPR} are two types of widely used recurrent units. 
% % (clear)
Using an RNN-based autoencoder for language representation learning was proposed by \citet{Dai2015SemisupervisedSL}. The model combines an LSTM encoder, and an LSTM decoder to learn language representation in an unsupervised fashion on the supervised evaluation datasets, and then finetunes the LSTM encoder for supervised tasks on the same datasets. They successfully show that learning the word representation and the compositionality of words could be done at the same time in an end-to-end machine learning system.

Since the RNN-based encoder processes an input sentence in the word order, it is obvious that the dependency of the representation on the starting words will decrease as the encoder processes more and more words. \citet{Tai2015ImprovedSR} modified the plain LSTM network to a tree-structured LSTM network, which helps the model to address the long-term dependency problem. Other than modifying the network structure, additional supervision could also help. \citet{Bowman2016AFU} proposed a model that learns to parse the sentence at the same time as the RNN is processing the input sentence. In the proposed model, the supervision comes from the objective function for the supervised tasks, and the parsed sentences, which means all training sentences need to be parsed prior to training. These two methods require additional preprocessing on the training data, which could be slow if we need to deal with a large corpus.

\begin{figure*}[th]
\centering
\includegraphics[width=0.8\textwidth]{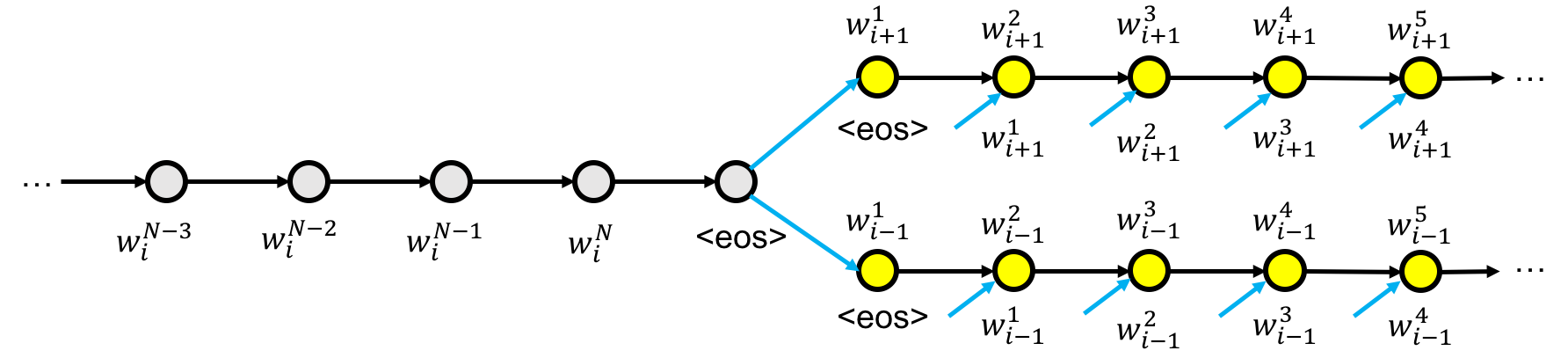}
\caption{The skip-thought neighbor model. The shared parameters are indicated in colors. The blue arrow represents the dependency on the representation produced from the encoder. For a given sentence $s_i$, the model tries to reconstruct its two neighbors, $s_{i-1}$ and $s_{i+1}$.}
\label{model}
\end{figure*}

Instead of learning to compose a sentence representation from the word representations, the skip-thought model \citet{Kiros2015SkipThoughtV} utilizes the structure and relationship of the adjacent sentences in the large unlabelled corpus. Inspired by the skip-gram model \cite{Mikolov2013EfficientEO}, and the sentence-level distributional hypothesis \cite{harris1954distributional}, the skip-thought model encodes the current sentence as a fixed-dimension vector, and instead of predicting the input sentence itself, the decoders predict the previous sentence and the next sentence independently. The skip-thought model provides an alternative way for unsupervised sentence representation learning, and has shown great success. The learned sentence representation encoder outperforms previous unsupervised pretrained models on 8 evaluation tasks with no finetuning, and the results are comparable to supervised trained models. In \citet{Triantafillou2016TowardsGS}, they finetuned the skip-thought models on the Stanford Natural Language Inference (SNLI) corpus \cite{Bowman2015ALA}, which shows that the skip-thought pretraining scheme is generalizable to other specific NLP tasks.

In \citet{Hill2016LearningDR}, the proposed FastSent model takes summation of the word representations to compose a sentence representation, and predicts the words in both the previous sentence and the next sentence. The results on this semantic related task is comparable with the RNN-based skip-thought model, while the skip-thought model still outperforms the FastSent model on the other six classification tasks. Later, Siamese CBOW \cite{Kenter2016SiameseCO} aimed to learn the word representations to make the cosine similarity of adjacent sentences in the representation space larger than that of sentences which are not adjacent.

Following the skip-thought model, we designed our skip-thought neighbor model by a simple modification. Section \ref{approach} presents the details.

\section{Approach}
\label{approach}
In this section, we present the skip-thought neighbor model. We first briefly introduce the skip-thought model \cite{Kiros2015SkipThoughtV}, and then discuss how to explicitly modify the decoders in the skip-thought model to get the skip-thought neighbor model.

\subsection{Skip-thought Model}
In the skip-thought model, given a sentence tuple $(s_{i-1},s_i,s_{i+1})$, the encoder computes a fixed-dimension vector as the representation $\mathbf{z}_i$ for the sentence $s_i$, which learns a distribution $p(\mathbf{z}_i|s_i;\theta_e)$, where $\theta_e$ stands for the set of parameters in the encoder. Then, conditioned on the representation $\mathbf{z}_i$, two separate decoders are applied to reconstruct the previous sentence $s_{i-1}$, and the next sentence $s_{i+1}$, respectively. We call them previous decoder $p(s_{i-1}|\mathbf{z}_i;\theta_{p})$ and next decoder $p(s_{i+1}|\mathbf{z}_i;\theta_{n})$, where $\theta_\cdot$ denotes the set of parameters in each decoder. 

Since the two conditional distributions  learned from the decoders are parameterized independently, they implicitly utilize the sentence order information within the sentence tuple. Intuitively, given the current sentence $s_i$, inferring the previous sentence $s_{i-1}$ is considered to be different from inferring the next sentence $s_{i+1}$.

\subsection{Encoder: GRU}
In order to make the comparison fair, we choose to use a recurrent neural network with the gated recurrent unit (GRU) \cite{Cho2014LearningPR}, which is the same recurrent unit used in \citet{Kiros2015SkipThoughtV}. Since the comparison among different recurrent units is not our main focus, we decided to use GRU, which is a fast and stable recurrent unit. In addition, \citet{Chung2014EmpiricalEO} shows that, on language modeling tasks, GRU performs as well as the long short-term memory (LSTM) \cite{Hochreiter1997LongSM}.

Suppose sentence $s_i$ contains $N$ words, which are $w_i^1,w_i^2,...,w_i^N$. At an arbitrary time step $t$, the encoder produces a hidden state $\mathbf{h}_i^t$, and we regard it as the representation for the previous subsequence through time $t$. At time $N$, the hidden state $\mathbf{h}_i^N$ represents the given sentence $s_i$, which is $\mathbf{z}_i$. The computation flow of the GRU in our experiments is shown below (omitting the subscript $i$) :
\begin{align}
\begin{bmatrix}
\mathbf{m}^t\\
\mathbf{r}^t
\end{bmatrix}&=\sigma\left(\mathbf{W}_h\mathbf{h}^{t-1}+\mathbf{W}_x\mathbf{x}^t\right) \\
\hat{\mathbf{h}}^t&=\tanh\left(\mathbf{W}\mathbf{x}^t+\mathbf{U}\left(\mathbf{r}^t\odot\mathbf{h}^{t-1}\right)\right)\\
\mathbf{h}^t&=(1-\mathbf{m}^t)\odot\mathbf{h}^{t-1}+\mathbf{m}^t\odot\mathbf{\hat{\mathbf{h}}}^t
\label{gru}
\end{align}
where $\mathbf{x}^t$ is the embedding for the word $w_i^t$, $\mathbf{W_\cdot}$ and $\mathbf{U}$ are the parameter matrices, and $\odot$ is the element-wise product.

\subsection{Decoder: Conditional GRU}
The decoder needs to reconstruct the previous sentence $s_{i-1}$ and the next sentence $s_{i+1}$ given the representation $\mathbf{z}_i$. Specifically, the decoder is a recurrent neural network with conditional GRU, and it takes the representation $\mathbf{z}_i$ as an additional input at each time step. 
% The computation flow is shown below (omitting the subscript $i$ ): \begin{align}
% \begin{bmatrix}
% \mathbf{m}^t\\
% \mathbf{r}^t
% \end{bmatrix}&=\sigma\left(\mathbf{W}_h\mathbf{h}^{t-1}+\mathbf{W}_x\mathbf{x}^t+\mathbf{W}_z\mathbf{z}_i\right) \\
% \hat{\mathbf{h}}^t&=\tanh\left(\mathbf{W}\mathbf{x}^t+\mathbf{U}\left(\mathbf{r}^t\odot\mathbf{h}^{t-1}\right)+\mathbf{U}_z\mathbf{z}_i\right)\\
% \mathbf{h}^t&=(1-\mathbf{m}^t)\odot\mathbf{h}^{t-1}+\mathbf{m}^t\odot\mathbf{\hat{\mathbf{h}}}^t
% \label{cgru}
% \end{align}
% where $\mathbf{x}^t$ is the embedding for the word $w_{i-1}^t$ or the embedding for the word $w_{i+1}^t$, depending on which one of the $s_{i-1}$ and $s_{i+1}$ is being reconstructed.

\begin{table*}[t]
\fontsize{9}{12}\selectfont
\begin{center}
\begin{tabular}{c | c c c c c | c | c c c }
\hline
\multirow{2}{*}{Model} & \multirow{2}{*}{MR} & \multirow{2}{*}{CR} & \multirow{2}{*}{SUBJ} & \multirow{2}{*}{MPQA} & \multirow{2}{*}{TREC} & \multirow{2}{*}{MSRP (Acc/F1)} & \multicolumn{3}{c}{SICK} \\ \cline{8-10}
 &&&&&&& $r$ & $\rho$ & MSE \\ 
\hline
\hline
uni-N-1200 & 71.5 & 78.4 & 90.1 & 83.4 & 85.2 & 72.1 / 81.7 & 0.8108 & 0.7382 & 0.3498 \\ 
bi-N-1200  & 71.9 & 79.3 & 90.8 & 85.2 & 88.8 & 72.8 / 81.5 & 0.8294 & 0.7594 & 0.3192 \\ 
combine-N-1200 & 73.6 & 80.2 &	91.4 & 85.7 & 89.0 & 73.5 / 82.1 & 0.8381 & 0.7721 & 0.3039 \\ 
\hline
uni-skip-1200 & 72.1 & 77.8 & 90.5 & 84.2 & 86.0 & 71.5 / 80.8 & 0.8196 & 0.7510 & 0.3350 \\ 
bi-skip-1200 & 72.9 &79.3 & 90.6 & 85.2 & 87.6 & 72.9 / 81.6 & 0.8264 & 0.7535 & 0.3237 \\ 
combine-skip-1200  & 74.0 & 80.4 & 91.5 & 86.1 & 87.8 & 73.7 / 81.7 & 0.8312 & 0.7610 & 0.3164 \\ 
\hline
\hline

uni-N-1200+AE & 70.8 & 75.9 & 90.4 & 82.7 & 87.2 & 73.3 / 81.7 & 0.8128 & 0.7450 & 0.3493 \\ 
bi-N-1200+AE  & 71.1 & 78.4 & 90.8 & 83.7 & \textbf{89.6} & 73.2 / 81.7 & 0.8198 & 0.7572 & 0.3389 \\ 
combine-N-1200+AE & 72.4 & 78.6 & 91.5 & 84.3 & 88.7 & \textbf{75.4} / \textbf{83.0} & 0.8347 & 0.7680 & 0.3135 \\ 
\hline
uni-skip-1200+AE & 68.4 & 76.5 & 89.4 & 81.4 & 81.6 & 72.2 / 81.4 & 0.7888 & 0.7224 & 0.3874 \\ 
bi-skip-1200+AE & 70.0 & 76.4 & 89.6 & 81.4 & 86.0 & 72.8 / 81.4 & 0.7908 & 0.7249 & 0.3824 \\ 
combine-skip-1200+AE  & 71.6 & 78.0 & 90.7 & 83.2 & 83.2 & 73.2 / 81.4 & 0.8086 & 0.7410 & 0.3562 \\ 
% \hline
% \hline
% uni-skip-2400$^*$  & 75.5 & 79.3 & 92.1 & 86.9 & 91.4 & 73.0 / 81.9 & 0.8477 & 0.7780 & 0.2872\\
% bi-skip-2400$^*$  & 73.9 & 77.9 & 92.5 & 83.3 & 89.4 & 71.2 / 81.2 & 0.8405 & 0.7696 & 0.2995\\
% combine-skip-2400$^*$ & \textbf{76.5} & 80.1 & \textbf{93.6} & \textbf{87.1} & \textbf{92.2} & 73.0 / \textbf{82.0} & \textbf{0.8584} & \textbf{0.7916} & \textbf{0.2687}\\
\hline
\hline
uni-N-next-1200 &	73.3 &	80.0 &	90.7 &	84.9 &	84.8 &	73.8 / 81.9 &	0.8207	& 0.7512	& 0.3330 \\
bi-N-next-1200 &	75.0 &	80.5 &	91.1 &	86.5 &	87.4 &	72.3 / 81.3 &	0.8271	& 0.7630 &	0.3223 \\
combine-N-next-1200 &	\textbf{75.8} &	\textbf{81.8} &	\textbf{91.9} &	\textbf{86.8} &	88.6 &	75.0 / 82.5 & \textbf{0.8396} & \textbf{0.7739} & \textbf{0.3013} \\
\hline
\end{tabular}
\end{center}
\caption{The model name is given by \emph{encoder type - model type - model size} with or without autoencoder branch \emph{+AE}. Bold numbers indicate the best results among all models. Without the autoencoder branch, our skip-thought neighbor models perform as well as the skip-thought models, and our ``next'' models slightly outperform the skip-thought models. However, with the autoencoder branch, our skip-thought neighbor models outperform the skip-thought models. }
% ``$^*$"s are the results reported by Kiros et al. \shortcite{Kiros2015SkipThoughtV}.}
\label{quantitative}
\end{table*}

\subsection{Skip-thought Neighbor Model}
Our hypothesis is that, even without the order information within a given sentence tuple, the skip-thought model should behave similarly in terms of the reconstruction error, and perform similarly on the evaluation tasks. To modify the skip-thought model, given $s_i$, we assume that inferring $s_{i-1}$ is the same as inferring $s_{i+1}$. If we define $\{s_{i-1}, s_{i+1}\}$ as the two neighbors of $s_i$, then the inferring process can be denoted as $s_j\sim p(s|\mathbf{z}_i;\theta_{d})$, for any $j$ in the neighborhood of $s_i$. The conditional distribution learned from the decoder is parameterized by $\theta_d$.

In experiments, we directly drop one of the two decoders, and use only one decoder to reconstruct the previous sentence $s_{i-1}$ and next sentence $s_{i+1}$ at the same time, given the representation $\mathbf{z}_i$ of $s_i$. Our skip-thought neighbor model can be considered as sharing the parameters between the previous decoder and the next decoder in the original skip-thought model. An illustration is shown in Figure \ref{model}.

% The encoder computes a representation $\mathbf{z}_i$ for $s_i$, and the decoder generates two sequences to match the two neighbors $s_{i-1}$ and $s_{i+1}$. Since $s_{i-1}$ and $s_{i+1}$ are independent given $\mathbf{z}_i$, we have \begin{align} 
% &p(s_{i+1},s_{i-1}|\mathbf{z}_i;\theta_{d})\nonumber \\
% =&p(s_{i-1}|\mathbf{z}_i;\theta_{d})p(s_{i+1}|\mathbf{z}_i;\theta_{d}).
% \end{align} 
The objective at each time step is defined as the log-likelihood of the predicted word given the previous words, which is
\begin{align}
&\ell^t_{i,j}(\theta_e,\theta_d) = \log p(w_j^t|w_j^{<t},\mathbf{z}_i;\theta_e,\theta_d) \\
&\max_{\theta_e,\theta_d}\sum_i\sum_{j\in {\{i-1,i+1\}}}\sum_t \ell^t_{i,j}(\theta_e,\theta_d) 
\end{align}
% \begin{align}
% \fontsize{9}{12}\selectfont
% \sum_{j\in {\{i-1,i+1\}}}\sum_t \log p(w_j^t|w_j^{<t},\mathbf{z}_i;\theta_e,\theta_d) \nonumber\\
% \end{align}
where $\theta_e$ is the set of parameters in the encoder, and $\theta_d$ is the set of parameters in the decoder. The loss function is summed across the whole training corpus.

\subsection{Skip-thought Neighbor with Autoencoder}
Previously, we defined $\{s_{i-1}, s_{i+1}\}$ as the two neighbors of $s_i$. In addition, we assume that $s_i$ could also be a neighbor of itself. Therefore, the neighborhood of $s_i$ becomes $\{s_{i-1},s_i,s_{i+1}\}$.  Inferring $s_j\sim p(s|\mathbf{z}_i;\theta_{d})$ for any $j$ in the neighborhood of $s_i$ then involves adding an autoencoder path into to our skip-thought neighbor model. In experiments, the decoder in the model is required to reconstruct all three sentences $\{s_{i-1}, s_i, s_{i+1}\}$ in the neighborhood of $s_i$ at the same time. The objective function becomes 
\begin{align}
\max_{\theta_e,\theta_d}\sum_i\sum_{j\in {\{i-1,i,i+1\}}}\sum_t l^t_{i,j}(\theta_e,\theta_d) 
\end{align}

Previously \citet{Hill2016LearningDR} tested adding an autoencoder path into their FastSent model. Their results show that, with the additional autoencoder path, the performance on the classification tasks slightly improved, while there was no significant performance gain or loss on the semantic relatedness task.

We tested both our skip-thought neighbor model and the original skip-thought model with the autoencoder path, respectively. The results are presented in Sections \ref{settings}, \ref{QE} and \ref{QI}.

\subsection{Skip-thought Neighbor with One Target}
In our skip-thought neighbor model, for a given sentence $s_i$, the decoder needs to reconstruct the sentences in its neighborhood $\{s_{i-1},s_{i+1}\}$, which are two targets. We denote the inference process as $s_i \rightarrow \{s_{i-1},s_{i+1}\}$. For the next sentence $s_{i+1}$, the inference process is $s_{i+1} \rightarrow \{s_i,s_{i+2}\}$. In other words, for a given sentence pair $\{s_i,s_{i+1}\}$, the inference process includes $s_i \rightarrow s_{i+1}$ and $s_{i+1} \rightarrow s_i$.

In our hypothesis, the model doesn't distinguish between the sentences in a neighborhood. In this case, an inference process that includes $s_i \rightarrow s_{i+1}$ and $s_{i+1} \rightarrow s_i$ is equivalent to an inference process with only one of them. Thus, we define a skip-thought neighbor model with only one target, and the target is always the next sentence. The objective becomes
\begin{align}
&\max_{\theta_e,\theta_d}\sum_i\sum_t \ell^t_{i,i+1}(\theta_e,\theta_d) 
\end{align}

\section{Experiment Settings}
\label{settings}
The large corpus that we used for unsupervised training is the BookCorpus dataset \cite{Zhu2015AligningBA}, which contains 74 million sentences from 7000 books in total. 

All of our experiments were conducted in Torch7 \cite{torch}. To make the comparison fair, we reimplemented the skip-thought model under the same settings, according to \citet{Kiros2015SkipThoughtV}, and the publicly available theano code\footnote{https://github.com/ryankiros/skip-thoughts}. We adopted the multi-GPU training scheme from the Facebook implementation of ResNet\footnote{https://github.com/facebook/fb.resnet.torch}.

We use the ADAM \cite{Kingma2014AdamAM} algorithm for optimization. Instead of applying the gradient clipping according to the norm of the gradient, which was used in \citet{Kiros2015SkipThoughtV}, we directly cut off the gradient to make it within $[-1,1]$ for stable training.

The dimension of the word embedding and the sentence representation are $620$ and $1200$. respectively. For the purpose of fast training, all the sentences were zero-padded or clipped to have the same length.

\section{Quantitative Evaluation}
\label{QE}
We compared our proposed skip-thought neighbor model with the skip-thought model on 7 evaluation tasks, which include semantic relatedness, paraphrase detection, question-type classification and 4 benchmark sentiment and subjective datasets. After unsupervised training on the BookCorpus dataset, we fix the parameters in the encoder, and apply it as a sentence representation extractor on the 7 tasks. 

For semantic relatedness, we use the SICK dataset \cite{Marelli2014ASC}, and we adopt the feature engineering idea proposed by \citet{Tai2015ImprovedSR}. For a given sentence pair, the encoder computes a pair of representations, denoted as $u$ and $v$, and the concatenation of the component-wise product $u\cdot v$ and the absolute difference $|u-v|$ is regarded as the feature for the given sentence pair. Then we train a logistic regression on top of the feature to predict the semantic relatedness score. The evaluation metrics are Pearson’s $r$, Spearman’s $\rho$, and mean squared error $MSE$.

The dataset we use for the paraphrase detection is the Microsoft Paraphrase Detection Corpus \cite{Dolan2004UnsupervisedCO}. We follow the same feature engineering idea from \citet{Tai2015ImprovedSR} to compute a single feature for each sentence pair. Then we train a logistic regression, and 10-fold cross validation is applied to find the optimal hyper-parameter settings.

The 5 classification tasks are question-type classification (TREC) \cite{Li2002LearningQC}, movie review sentiment (MR) \cite{Pang2005SeeingSE}, customer product reviews (CR) \cite{Hu2004MiningAS}, subjectivity/objectivity classification (SUBJ) \cite{Pang2004ASE}, and opinion polarity (MPQA) \cite{Wiebe2005AnnotatingEO}.

In order to deal with more words besides the words used for training, the same word expansion method, which was introduced by \citet{Kiros2015SkipThoughtV}, is applied after training on the BookCorpus dataset. 

The results are shown in Table \ref{quantitative}, where the model name is given by \emph{encoder type - model type - model size}. We tried with three different types of the encoder, denoted as \emph{uni-}, \emph{bi-}, and \emph{combine-} in Table \ref{quantitative}. The first one is a uni-directional GRU, which computes a 1200-dimension vector as the sentence representation. The second one is a bi-directional GRU, which computes a 600-dimension vector for each direction, and then the two vectors are concatenated to serve as the sentence representation. Third, after training the uni-directional model and the bi-directional model, the representation from both models are concatenated together to represent the sentence, denoted as \emph{combine-}. 

In Table \ref{quantitative}, \emph{-N-} refers to our skip-thought neighbor model, \emph{-N-next-} refers to our skip-thought neighbor with only predicting the next sentence, and \emph{-skip-} refers to the original skip-thought model.

\subsection{Skip-thought Neighbor vs. Skip-thought}
From the results we show in Table \ref{quantitative}, we can tell that our skip-thought neighbor models perform as well as skip-thought models but with fewer parameters, which means that the neighborhood information is effective in terms of helping the model capture sentential contextual information. 

% As we expected, the skip-thought models with only one target perform as well as the skip-thought models on TREC and MPQA, and slightly better than skip-thought models on other 5 tasks. 

\subsection{Skip-thought Neighbor+AE vs. Skip-thought+AE}

For our skip-thought neighbor model, incorporating an antoencoder (+AE) means that, besides reconstructing the two neighbors $s_{i-1}$ and $s_{i+1}$, the decoder also needs to reconstruct $s_i$. For the skip-thought model, since the implicit hypothesis in the model is that different decoders learn different conditional distributions, we add another decoder in the skip-thought model to reconstruct the sentence $s_i$. The results are also shown in Table \ref{quantitative}.

As we can see, our skip-thought neighbor+AE models outperform skip-thought+AE models significantly. Specifically, in skip-thought model, adding an autoencoder branch hurts the performance on SICK, MR, CR, SUBJ and MPQA dataset. We find that the reconstruction error on the autoencoder branch decreases drastically during training, while the sum of the reconstruction errors on the previous decoder and next decoder fluctuates widely, and is larger than that in the model without the autoencoder branch. It seems that, the autoencoder branch hurt the skip-thought model to capture sentential contextual information from the surrounding sentences. One could vary the weights on the three independent branches to get better results, but it is not our main focus in this paper.

In our skip-thought neighbor model, the inclusion of the autoencoder constraint did not have the same problem. With the autoencoder branch, the model gets lower errors on reconstructing all three sentences. However, it doesn't help the model to perform better on the evaluation tasks.

\subsection{Increasing the Number of Neighbors}
We also explored adding more neighbors into our skip-thought neighbor model. Besides using one decoder to predict the previous 1 sentence, and the next 1 sentence, we expand the neighborhood to contain 4 sentences, which are the previous 2 sentences, and the next 2 sentences. In this case, the decoder is required to reconstruct 4 sentences at the same time. We ran experiments with our model, and we evaluated the trained encoder on 7 tasks. 

There is no significant performance gain or loss on our model trained with 4 neighbors; it seems that, increasing the number of neighbors doesn't improve the performance, but it also doesn't hurt the performance. Our hypothesis is that, reconstructing four different sentences in a neighborhood with only one set of parameters is a hard task, which might distract the model from capturing the sentential contextual information.

\subsection{Skip-thought Neighbor with One Target}
Compared to the skip-thought model, our skip-thought neighbor model with  one target contains fewer parameters, and runs faster during training, since for a given sentence, our model only needs to reconstruct its next sentence while the skip-thought model needs to reconstruct its surrounding two sentences. The third section in Table \ref{quantitative} presents the results of our model with only one target. Surprisingly, it overall performs as well as the skip-thought models as all previous models.

\subsection{A Note on Normalizing the Representation}
An interesting observation was found when we were investigating the publicly available code for \citet{Kiros2015SkipThoughtV}, which is, during training, the representation produced from the encoder will be directly sent to the two decoders, however, after training, the output from the encoder will be normalized to keep the l2-norm as 1, so the sentence representation is a normalized vector.

\begin{table}[h]
\fontsize{9}{12}\selectfont
\begin{center}
\begin{tabular}{c | c c c}
\hline
\multirow{2}{*}{Model} &  \multicolumn{3}{c}{SICK} \\ \cline{2-4}
& $r$ & $\rho$ & MSE \\ 
\hline
\hline
uni-N-1200 & 0.8174 & 0.7409 & 0.3404 \\
bi-N-1200 & 0.8339 & 0.7603 & 0.3102 \\
combine-N-1200 & 0.8360 & 0.7645 & 0.3054 \\
\hline
uni-skip-1200 & 0.8232 & 0.7493 & 0.3292 \\
bi-skip-1200 & 0.8280 & 0.7526 & 0.3203 \\
combine-skip-1200 & 0.8340 & 0.7600 & 0.3100 \\
\hline
\hline
uni-N-1200+AE & 0.8210 & 0.7450 & 0.3493 \\
bi-N-1200+AE & 0.8302 & 0.7610 & 0.3169 \\
combine-N-1200+AE & 0.8326 & 0.7621 & 0.3285 \\
\hline
uni-skip-1200+AE & 0.7959 & 0.7255 & 0.3760 \\
bi-skip-1200+AE & 0.7986 & 0.7282 & 0.3693 \\
combine-skip-1200+AE & 0.8129 & 0.7400 & 0.3508 \\
\hline
\hline
uni-N-next-1200 & 0.8253 & 0.7508 & 0.3245 \\
bi-N-next-1200 & 0.8296 & 0.7621 & 0.3175 \\
combine-N-next-1200 & \textbf{0.8402} & \textbf{0.7706} & \textbf{0.2999} \\
\hline
\end{tabular}
\end{center}
\caption{Evaluation on SICK dataset without normalizing the representation. Bold numbers indicate the best values among all models.}
\label{unnormalize}
\end{table}

We conducted experiments on the effect of normalization during the evaluation, and we evaluated both on our skip-thought neighbor model, and our implemented skip-thought model. Generally, the normalization step slightly hurts the performance on the semantic relatedness SICK task, while it improves the performance across all the other classification tasks. The Table \ref{quantitative} presents the results with the normalization step, and Table \ref{unnormalize} presents the results without normalization on SICK dataset.

\begin{table*}[!ht]
\fontsize{10}{13}\selectfont
\begin{center}
\begin{tabular}{l}
\hline
\hline
\textbf{i wish i had a better answer to that question .}\\
i wish i knew the answer .\\
i only wish i had an answer .\\
\hline
\textbf{i kept my eyes on the shadowed road , watching my every step .}\\
i kept my eyes at my feet and up ahead on the trail .\\
i kept on walking , examining what i could out of the corner of my eye .\\
\hline
\textbf{the world prepared to go into its hours of unreal silence that made it seem magical , and it really was .}\\
the world changed , and a single moment of time was filled with an hour of thought .\\
everything that was magical was just a way of describing the world in words it could n't ignore .\\
\hline
\textbf{my phone buzzed and i awoke from my trance .}\\
i flipped my phone shut and drifted off to sleep .\\
i grabbed my phone and with groggy eyes , shut off the alarm .\\
\hline
\textbf{i threw my bag on my bed and took off my shoes .}\\
i sat down on my own bed and kicked off my shoes . \\
i fell in bed without bothering to remove my shoes . \\
\hline
\hline
\end{tabular}
\end{center}
\caption{In each section, the first sentence is the query, the second one is the nearest neighbor retrieved from the database, and the third one is the 2nd nearest neighbor. The similarity between every sentence pair is measure by the cosine similarity in the representation space.}
\label{sr}
\end{table*}

\section{Qualitative Investigation}
We conducted investigation on the decoder in our trained skip-thought neighbor model.
\label{QI}
% \subsection{Training Procedure}
% We plot the reconstruction error during the training procedure to compare the difference between our skip-thought neighbor model and skip-thought model, which can be found in Figure \ref{loss}. To make sure the comparison is fair, we ensure that the data in each batch is the same for training our skip-thought neighbor model and skip-thought model. 

% Surprisingly, the skip-thought neighbor model behaves similarly with the skip-thought model in terms of the training error, and our skip-thought neighbor model gets even slightly lower training error than the skip-thought model, which verifies our hypothesis that two decoders in the skip-thought model are learning general neighborhood information, not the order information that the model wants to learn.

% \begin{figure}
% \begin{center}
% \includegraphics[width=0.45\textwidth]{figs/loss}
% \end{center}
% \caption{The figure shows the variation of the reconstruction error during training for one epoch, and we compare 4 models, which are listed in the legend. With the same hyperparameter settings, bi-skip-thought neighbor model behaves similar to bi-skip model.}
% \label{loss}
% \end{figure}
\subsection{Sentence Retrieval}
We first pick up 1000 sentences as the query set, and then randomly pick up 1 million sentences as the database. In the previous section, we mentioned that normalization improves the performance of the model, so the distance measure we applied in the sentence retrieval experiment is the cosine distance. Most of retrieved sentences look semantically related and can be viewed as the sentential contextual extension to the query sentences. Several samples can be found in Table \ref{sr}. 

\subsection{Conditional Sentence Generation}
\begin{table}[!h]
\fontsize{10}{12}\selectfont
\begin{center}
\begin{tabular}{l}
\hline
\hline
`` i do n't want to talk about it . '' \\
`` i 'm not going to let you drive me home . '' \\
`` hey , what 's wrong ? '' \\
i 'm not sure how i feel about him . \\
i was n't going to be able to get to the beach . \\
he was n't even looking at her .\\
`` i guess you 're right . '' \\
\hline
\hline
\end{tabular}
\end{center}
\caption{Samples of the generated sentences.}
\label{sents}
\end{table}

Since the models are trained to minimizing the reconstruction error across the whole training corpus, it is reasonable to analyze the behavior of the decoder on the conditional sentence generation. We first randomly pick up sentences from the training corpus, and compute a representation for each of them. Then, we greedily decode the representations to sentences. Table \ref{sents} presents the generated sentences. Several interesting observations worth mentioning here.

The decoder in our skip-thought neighbor model aims to minimize the distance of the generated sentence to two targets, which lead us to doubt if the decoder is able to generate at least grammatically-correct English sentences. But, the results shows that the generated sentences are both grammatically-correct and generally meaningful.

We also observe that, the generated sentences tend to have similar starting words, and usually have negative expression, such as \emph{i was n't}, \emph{i 'm not}, \emph{i do n't}, etc. After investigating the training corpus, we noticed that this observation is caused by the dataset bias. A majority of training sentences start with \emph{i} and \emph{i 'm} and \emph{i was }, and there is a high chance that the negation comes after \emph{was} and \emph{'m}. In addition, the generated sentences rarely are the sentential contextual extension of their associated input sentences, which is same for the skip-thought models. More investigations are needed for the conditional sentence generation.

\section{Conclusion}
We proposed a hypothesis that the neighborhood information is effective in learning sentence representation, and empirically tested our hypothesis. Our skip-thought neighbor models were trained in an unsupervised fashion, and evaluated on 7 tasks. The results showed that our models perform as well as the skip-thought models. Furthermore, our model with only one target performs better than the skip-thought model. Future work could explore more on the our skip-thought neighbor model with only one target, and see if the proposed model is able to generalize to even larger corpora, or another corpus that is not derived from books.

\section*{Acknowledgments}
We gratefully thank Jeffrey L. Elman, Benjamin K. Bergen, Seana Coulson, and Marta Kutas for insightful discussion, and thank Thomas Andy Keller, Thomas Donoghue, Larry Muhlstein, and Reina Mizrahi for suggestive chatting. We also thank Adobe Research Lab for GPUs support, and thank NVIDIA for DGX-1 trial as well as support from NSF IIS 1528214 and NSF SMA 1041755.

% \vspace{-0.6cm}
\bibliography{acl2017}
\bibliographystyle{acl_natbib}

\end{document}